\documentclass[fleqn,10pt]{SelfArx} 

\usepackage{amsmath,amssymb,amsfonts}
\usepackage{algorithmic}
\usepackage{graphicx}
\usepackage{textcomp}
\usepackage[hyphens]{url}
\usepackage{multirow}
\usepackage{xcolor}
\usepackage{authblk}
\usepackage{float}
\usepackage[numbers]{natbib}

\definecolor{color1}{RGB}{0,0,90} 
\definecolor{color2}{RGB}{0,20,20} 

\usepackage{hyperref} 
\hypersetup{hidelinks,colorlinks,breaklinks=true,urlcolor=color2,citecolor=color1,linkcolor=color1,bookmarksopen=false,pdftitle={Title},pdfauthor={Author}}

\JournalInfo{Draft Version} 
\Archive{} 

\PaperTitle{Interruptions detection in video conferences} 

\Authors{Shmuel Horowitz, Dima Kagan, Galit Fuhrmann Alpert, and Michael Fire} 
\affiliation{\textsuperscript{1}\textit{\href{http://data4good.io}{The Data4Good Lab}, Department of Software and Information Systems Engineering, Ben-Gurion University of the Negev, Israel}} 
\affiliation{shmuelho@post.bgu.ac.il, kagandi@post.bgu.ac.il, fuhrmann@bgu.ac.il, mickyfi@bgu.ac.il} 

\newcommand{\keywordname}{Keywords} 

\Abstract{In recent years, video conferencing (VC) popularity has skyrocketed for a wide range of activities. As a result, the number of VC users surged sharply. 
The sharp increase in VC usage has been accompanied by various newly emerging privacy and security challenges. VC meetings became a target for various security attacks, such as Zoombombing. Other VC-related challenges also emerged. For example, during COVID lockdowns, educators had to teach in online environments struggling with keeping students engaged for extended periods. In parallel, the amount of available VC videos has grown exponentially. Thus, users and companies are limited in finding abnormal segments in VC meetings within the converging volumes of data. Such abnormal events that affect most meeting participants may be indicators of interesting points in time, including security attacks or other changes in meeting climate, like someone joining a meeting or sharing a dramatic content. 

Here, we present a novel algorithm for detecting abnormal events in VC data. We curated VC publicly available recordings, including meetings with interruptions. We analyzed the videos using our algorithm, extracting time windows where abnormal occurrences were detected. 
Our algorithm is a pipeline that combines multiple methods in several steps to detect users’ faces in each video frame, track face locations during the meeting and generate vector representations of a facial expression for each face in each frame. Vector representations are used to monitor changes in facial expressions throughout the meeting for each participant. The overall change in meeting climate is quantified using those parameters across all participants, and translating them into event anomaly detection. This is the first open pipeline for automatically detecting anomaly events in VC meetings. Our model detects abnormal events with 92.3\% precision over the collected dataset.}

\begin{document}

\flushbottom 

\maketitle 

\thispagestyle{empty} 


\section{Introduction} 

\addcontentsline{toc}{section}{Introduction} 

Video conferencing (VC) technology has been available for several decades. However, as a result of the COVID-19 pandemic outbreak, people discovered that video conferencing could replace face-to-face meetings for a wide range of in-person activities, including for example, school, work, and other social meetings. As a result, the number of video conferencing users, along with the number of daily  meetings, exhibited a sharp surge \cite{DBLP:journals/corr/abs-2007-01059}. Today, hundreds of millions of video conferencing meetings take place daily, encompassing millions of users \cite{DBLP:journals/corr/abs-2007-01059}. Moreover, a recent study  \cite{AmericanWorkforce2025} estimates that by 2025,  22\% of the American workforce will be working remotely, indicating that video conferencing is here to stay.
The sharp increase in video conferencing usage has been accompanied by various newly emerging challenges. For example, during covid lockdowns, educators had to teach in online environments struggling with keeping remote students engaged for extended periods of time. 

Over the past two years, the amount of available video conference video has grown exponentially. However, users and companies are limited in their ability to find abnormal segments in video meetings within the converging volumes of data.
Such abnormal segments could indicate the most interesting moment in a meeting or lecture or even an abnormal event that should not have occurred in the first place.
For example, the rise in popularity of video conferencing brought a new type of online harassment (also referred to as Zoomboming), where an uninvited person interrupts a video meeting by joining and sharing inappropriate content~\cite{SPACE_INVADERS}, such as racist~\cite{article_Zoom-ing-on-White-Supremacy}, misogynist attacks~\cite{article_Zoom-ing-misogynist} or sexual harassment~\cite{sexual_harassment_vc}.

\begin{figure*}[h]
\includegraphics[width=0.8\textwidth]{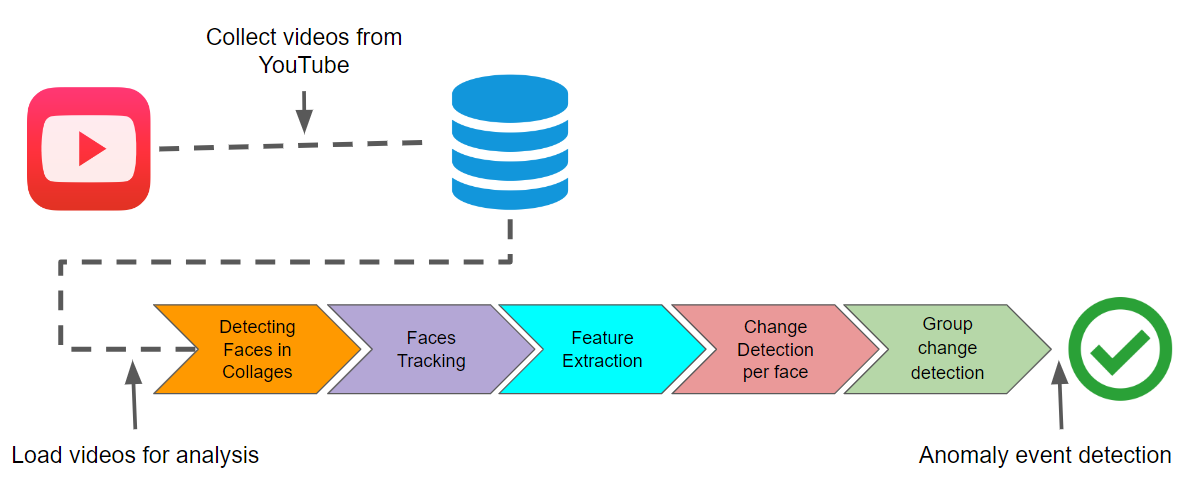}
\centering
\caption{Full algorithm pipeline: from the original video to abnormal event detection.}
\label{fig:method's steps}
\end{figure*}

While to date, there are studies that address anomaly detection in videos \cite{1902.05872}\cite{https://doi.org/10.48550/arxiv.2004.05993}\cite{DBLP:journals/corr/abs-2004-00222}, they are mostly based on open street scenarios based on well-known datasets.
This makes these methods irrelevant to this new type of video, where each participant has his own area in the video with a separate stream.
In video conferencing videos, the abnormal events are reflected in the faces of the meeting participants themselves rather than in the surrounding scenes.

In this study, we present a novel algorithm for detecting abnormal events in video conferencing videos. We consider an abnormal event as an event that affects most of the meeting participants. Abnormal events include Zoombombing attacks, but also normal changes in meeting climate - such as a teacher entering an online classroom, or even someone telling a funny joke. To this end, we curated and annotated video conferencing recordings, including meetings with Zoombombing events. We analyzed the curated videos using our algorithm and we show that we were able to extract the annotated moments where abnormal occurrences were observed. 

Our algorithm is based on combining multiple methods to extract face-related features about  meeting participants (see Figure~\ref{fig:method's steps}).

The meeting participants' faces are detected in every frame. Face locations are tracked during the meeting, and an expression representation is generated for each face at a frame rate of 4 fps to monitor their expressions throughout the meeting. We subsequently quantify the change in facial expression for each participant, and these parameters are used to quantify the overall change in the meeting to pursue anomaly detection further.

To the best of our knowledge, this paper presents the first available pipeline for detecting anomaly events in video conferences.
We published the implementation as an open-code framework (see Section \ref{opencode_and_dataset}). Furthermore, we offer our in-house curated and annotated  first-of-its-kind dataset of VC recordings containing annotated data of anomalies and unusual events with a total length of over eight hours. We show that our model detects abnormal events with a precision of 92.3\% and recall of 72.5\% over the collected data set.
As a bonus, we also demonstrated that the algorithm could additionally monitor video conference participants' engagement levels. This is particularly important and potentially applicable for education purposes, where student engagement influences multiple factors, including student satisfaction and desire to learn and the teacher's motivation.  

The remainder of the paper is organized as follows: In Section~\ref{related_work}, we present an overview of relevant studies. In Section~\ref{methods}, we describe the methods we used. In Section~\ref{experiments_results} we describe the datasets,  which we collected and annotated, the experimental setup, the evaluation metrics used throughout this study and the results of our analysis using suggested our framework. In Section~\ref{discussion}, we discuss the results, practical implications, and limitations of the work. Lastly, in Section~\ref{conclusions}, we present our conclusions from this study.

\section{Related Work}\label{related_work}
This work combines several fields of research: Facial expressions recognition, anomaly detection in time series and anomaly detection in videos. In the following subsections, we will review related studies in these fields.

\subsection{Facial Expression Recognition}
Facial expression recognition (FER) is a research area of broad impact on multiple domains, including intelligent security, entertainment, virtual reality, education, etc.  Seven universal facial expressions are standardly defined in the professional literature \cite{Ekman1994StrongEF}: Happiness, Anger, Sadness, Disgust, Surprise, Fear, and Neutral expressions. Therefore, datasets in this research field  contain mostly images of these annotated categories, and most research papers present models developed specifically to recognize these predefined facial expression categories. 
The Fer-2013 dataset\footnote{https://www.kaggle.com/c/challenges-in-representation-learning-facial-expression-recognition-challenge} is the most commonly used image dataset in this field, launched on Kaggle as part of the ICML 2013  recognition challenge. The dataset contains 28709 training images, 3589 validation images, and 3589 test images, split into those seven categories. There are also additional datasets, like CK+~\cite{inproceedings_ck} and SFEW~\cite{1013bbb172214ce5ad895e9a31e0865b}, but they are much smaller and simple for recognition tasks.
In this study, we focus our analysis on the first dataset - FER2013 - as it is a much larger and more diverse dataset. Models trained on this dataset show improved performance in extracting facial features and less over-fitting to the training data.

Various FER models have been developed over the years .
In recent years, as neural networks have evolved with the field of computer vision in general, they also invaded the FER area~\cite{revina2021survey}. In 2016, Pramerdorfer and Kampel~\cite{Pramerdorfer2016FacialER} developed a five-layer convolutional neural network (CNN), which was trained on the FER-2013 and  CK+ datasets, to classify images into one of the seven different expressions,  achieving an accuracy of  61\%. In their study, they also used VGG architecture, inception net, and Resnet and improved the accuracy results to 72.7\%, 71.4\%, and 72.4\%, respectively \cite{Pramerdorfer2016FacialER}. By using an ensemble of those CNN’s the classification accuracy was improved to 75.2\%.
The best-known model to address this challenge was presented in 2021 by Momin et al.~\cite{momin2021recognizing}. Momin et al. developed an ensemble learning technique for processing convoluted facial representations using a multi-layer perceptron as a meta-classifier. They used an ensemble of feature-extracting neural networks (backbones) to encode the network's input into a certain feature representation, such as sequential block, dense block, and others, and achieved an accuracy of 85.07\% on FER-2013 dataset  \cite{momin2021recognizing}.

\subsection{Time-series Anomaly Detection}
Time-series analysis has become important in diverse fields, including medicine, finance, business, cybersecurity, aerospace, etc. Time series data are sequences of measurements that are collected over time in order to capture  the time-dependent behavior of systems. Outliers in time series data are values that significantly differ from the patterns and trends of the other values in the time series. Anomaly detection problems in time series could be formulated to find outlier data points that deviate significantly from most observations. 
Anomaly detection methods on time-series data can be divided into three main categories \cite{DBLP:journals/corr/abs-2004-00433}:
(a) statistical methods;  (b) classical machine learning methods; and (c) deep learning based methods.

\textit{Statistical approaches} are based on the assumption that observed data is generated by a specific underlying statistical model \cite{breiman_statistical_2001} and the analysis attempts to identify a sample that exceeds the normal behavior of the underlying model. Many statistical-based algorithms are aimed at static data, assuming that its mean and variance are constant over time. The best-known methods in this category are Auto-Regression (AR), and the Moving Average (MA) models \cite{inbook}, which can be used together as an Autoregressive Moving Average (ARMA) model. The AR model assumes a dependent linear relation between the observation and the values of a specified number of previous observations plus an error term.  The Moving Average (MA) model accounts for the dependency between an observation and the residual errors by applying a moving average model to previous observations. The ARMA  model \cite{breiman_statistical_2001} integrates these two models, where ARIMA and SARIMA models \cite{7814437} as a generalization of the ARMA model. The ARIMA makes the ARMA suitable for non-stationary time series, and the SARIMA makes the ARIMA suitable for seasonal time series. 

The second category is \textit{machine learning} based appraoches \cite{DBLP:journals/corr/abs-2004-00433}. These algorithms aim at detect time-series anomalies without assuming a specific underlying generative or statistical model for data generation. Chen and Fei Li \cite{CHEN2011178} showed how to use the Density-Based Spatial Clustering of Application with Noise algorithm (DBSCAN) to detect outliers in the data. Using this method, they identify anomalies as small and far clusters created by the DBSCAN algorithm.  Another approach was shown by Liu et al. \cite{10.1109/ICDM.2008.17}, where an ensemble of Isolation Trees (iTrees)  is used to describe the data.  According to Liu et al.~\cite{10.1109/ICDM.2008.17}, it is more likely that anomalies are closer to the root of an iTree because they are more likely to be isolated than non-anomalous points.
A Support Vector Machine (SVM) based model was described by Scholkopf et al. \cite{10.5555/3009657.3009740} as a semi-supervised approach where the training set consists of only one class of the normal data. Time-series data is projected into a set of vectors that the SVM classifier tries to separate normal and abnormal samples. Extreme Gradient Boosting decision trees algorithm (GBM and XGBoost)~\cite{DBLP:journals/corr/ChenG16} also could assist with time series anomaly detection by predicting the next sample and computing the error between the prediction and the true sample \cite{Henriques2020CombiningKA}. The distribution of this error and mark extreme errors can be computed in terms of variance as anomaly samples.

The third category is \textit{deep learning and neural networks} based models. Munir et al. \cite{8581424} used CNN architecture to forecast the following sample in time-series and detect anomalies based on the error of the prediction. Another approach has been suggested by Zheng et al.~\cite{10.1007/978-3-319-08010-9_33}. They used CNN to classify time series samples in a supervised dataset with one class of normal data to detect the anomalies. Zheng et al.~\cite{7344872} used Long Short-Term Memory networks (LSTM) to predict time series samples. Timestamps were marked as normal or abnormal samples according to the prediction error of the sample~\cite{7344872}.

\subsection{Abnormal Event Detection in Videos}
Anomaly detection in videos suffers from an additional challenge, as the definition of an anomalous event is ambiguous and depends on the context~\cite{8578776}. For example, a running person could be a regular event in the park but an unusual event when it takes place inside a restaurant, and an incident of people laughing would be considered a regular incident in a stand-up show but an unusual incident during a lecture in which the participants suddenly start laughing.
Another difficulty stems from a lack of benchmark datasets similar to  those that are available for other computer vision challenges. The two primary datasets in this domain are UBI-Fights \cite{degardin2020human} and ucsd-ped2~\cite{Mahadevan2010AnomalyDI}. The UBI-Fights dataset contains 216 videos of fight events and 784 videos of normal daily life situations. The Ucsd-ped2 dataset was acquired using a stationary camera mounted at an elevation overlooking pedestrian walkways. Normal events contain only pedestrians, and abnormal events contain non-pedestrians entities in the walkways or anomalous pedestrian motion patterns. 
Many research papers have dealt with these issues in different ways. However, the focus was on open scenes, as available in the two datasets mentioned above.  
Several approaches tried to automatically learn motion patterns from the data through histogram-based methods \cite{1bef15bddc56496a955930bc3b5c8d4d}, topic modeling~\cite{Hospedales2009AMC}, mixtures of dynamic textures model~\cite{6531615}, local spatio-temporal motion pattern models~\cite{5206771}, Spatiotemporal Autoencoder~\cite{Chong2017AbnormalED}, and multiple instance learning (MIL) \cite{8578776}. Combinations of Convolutional neural network (CNN) and LSTM were used by Vignesh et al.~\cite{8015002}, Sharma et al.~\cite{sharma2021efficient} and Ullah et al. \cite{ullah2021cnn}.
Anomaly detection in video sequence was conducted by learning a correspondence between common object appearances and their associated motions, demonstrated by Nguyen et al.~\cite{Nguyen_2019_ICCV}. Other methods for anomalies detection  based on self-supervised and semi-supervised methods which were applied in order to tackle the lack of annotated datasets were suggested by Wu et al.~\cite{wu2020not}, Zhang et al. \cite{zhang2019temporal} and Liu et al.~\cite{liu2019margin}.

\section{Methods}\label{methods}
The main goal of this study was to develop a method for detecting abnormal events in video conferencing meetings.
To address this goal, we  developed a generic pipeline that consists of several stages, as illustrated in Figure~\ref{fig:method's steps}. Notably, the pipeline is modular, and each step can be implemented using various algorithms. For example,  other models could be considered for face detection, extracting features from participants' behavior, detecting anomalies in time series, etc. 
The methods we used for each pipeline stage are described below.

 \subsection{Detecting Faces in Collages of Participants} \label{detection_method}
In the first step of the pipeline, we detect the face of the participants in the conversation.
Our algorithm accepts an offline record of a video conference as input. 
We apply a combination of two existing face detection tools to detect faces in the zoom collage image. The first tool is a pre-trained YOLO 3~\cite{redmon2018yolov3} model,\footnote{https://github.com/OlafenwaMoses/ImageAI/tree/master/imageai/Detection} and the second tool is a RetinaFaceNet with ResNet-50 net as a backbone~\cite{DBLP:journals/corr/abs-1905-00641}.
 By utilizing both these two tools, we extract
the bounding boxes of the faces in each image collage. The output bounding boxes of the two algorithms are unified by determining the overlap area between them and a heuristic that optimizes the final detection. 
Given two bounding boxes $a$ and $b$ generated by two separate models, we strive to determine if $a$ and $b$  represent two different faces or the same one. 
We formally define $f(a,b)$ as the merge function of the two bounding boxes $a$ and $b$ as:
\[
    f(a,b)= 
\begin{cases}
    a,& \text{if } a \cap b\ = a\\
    b,& \text{if } a \cap b\ = b\\
     a \cap b,& \text{if } IOU(a,b)\geq threshold\\
    \{a,b\},              & \text{otherwise}
\end{cases}
\]
In the case where one of the detection bounding boxes fully contained the second detection bounding box ($a \cap b\ = a$ or $a \cap b\ = b$), we consider the inner bounding box. Otherwise, we take the intersection area between them as the bounding box to consider if the  Intersection over Union (IOU) between the two bounding boxes is greater than a set $threshold$. The threshold value was chosen empirically by examining images in which there was a miss by one of the detectos. We examined at what threshold we would get 95\% of the faces in at least one of the detectors. In any other case, we consider the detections as two separate detections. 

 \subsection{Faces Tracking} 
The second step in the pipeline is designed to track each participant's face in the video stream across time in order to account for cases where the participant's location changes from frame to frame. 
Most commonly, during video conference meetings, each face remains in the same location within the multi-face grid image.
This means that the IOU between the face's position in sequence frames should have a value near 1. 
However, after manually inspecting dozens of video conference recordings, we noticed that in some cases, the locations of faces within the faces grid change or get shuffled.
This can happen for several reasons, including when a participant enters
or leaves the meeting,  when the number of participants changes, or  when there is a change of speaker. To associate a participant’s face in consecutive frames, we thus utilized the bounding box achieved in the previous stage and measured the IOU between each pair of faces in the two consecutive frames. If all pairs in the current and previous image had IOU greater than a threshold, we associated the face pairs with each other. In any other case, we linked the current frame faces to the last frame faces by comparing the detected faces in each frame to the faces in the previous frame with the face recognition tool. 
To this end, we used DeepFace model,\footnote{\url{https://github.com/serengil/deepface}} which is a lightweight face recognition and facial attribute analysis open-sourced library for Python, which makes it easy and quick to run the model on large volumes of videos. 
Since this tool is sensitive to face resolution in case of a dense grid and small size participants' images or to hide dynamic changes in the participant's appearance - we did not use it in all cases and only as a second step. 

\subsection{Features Extraction}
The third step of the pipeline is designed to extract image features that can indicate a change in participants' behavior and can potentially be related to an abnormal conversation event. To examine feature changes in the facial image, we chose to apply and compare three different approaches for feature extraction.

The first approach (referred to as  128-dimensional embedding) is based on a 128-dimensional embedding vector representation of a face, as obtained with another face recognition tool based on ResNet-50 net as a backbone,\footnote{\url{https://github.com/ageitgey/face_recognition_models}} This tool is a well known open-source package for face recognition that also enables obtaining the face's embedding vector. The embedding vector is a projection of the face image onto a small Euclidean space, where two embedding vectors of similar faces would be closer to each other in the Euclidean space. We tested the hypothesis that similar expressions of the same individual can also be reflected in the embedding vector, such that major changes in facial expressions would be represented by feature vectors that are far apart in the Euclidean space.  

The second approach (referred to as 7-dimensional vector) is mapping facial expressions onto a representative vector. We attempted to encode the face into a latent space that is more meaningful in terms of the participant’s response to special events.
For training this encoder, we first trained a network for the classification of facial expressions. We utilized the FER2013  dataset containing 28709 training images, 3589 validation images, and 3589 test images.\footnote{\url{https://www.kaggle.com/c/challenges-in-representation-learning-facial-expression-recognition-challenge}} The image size is 48x48. Facial expression is classified into seven different types. We used the five layers CNN model demonstrated by Dufourq and Bassett \cite{dufourq2020evolutionary} to train a classification model where the model's output was one of the seven expression class. To obtain the encoder, we took the classification model's last fully connected layer output, a seven-dimension vector representing the expression’s latent space, and used this vector of expressions to examine the participant's facial expressions change on our own data.
It is worth noting that although the SOTA model gives 85\% accuracy \cite{momin2021recognizing}, we used an inferior model in terms of accuracy because our emphasis was on using the output of the last layer and not its final classification.

The third approach (referred to as deep-face expression) we use is the facial expression classification \label{third_approach_detection}(one of the seven basic expressions) as a  simple measure to show what the participant is going through. In other words: We take into account only the most dominant expression in the emotion embedding vector. To this end, we used the mentioned above DeepFace model, which can also classify the expression, age, and gender of a face.
 
 \subsection{Change Detection}
 \label{change_detection_section} 
\begin{figure}[!h]
\includegraphics[scale=0.4]{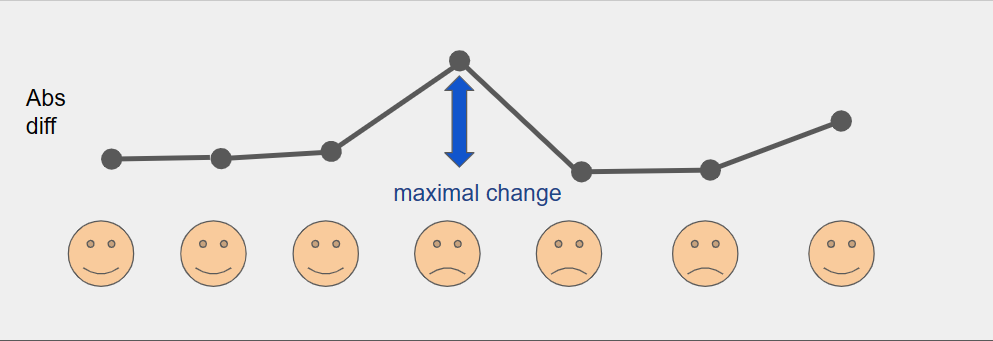}
\centering
\caption{Maximal change between faces is expected when a drastic change between expressions occurs over time.}
\label{fig:maxdf}
\end{figure}

 In the fourth step, we use the extracted facial features and analyze changes in those features over time for each participant separately. We use anomaly detection in time series methods to detect significant changes in facial expressions at different time frames within a video. We assume that small changes in facial expressions occur regularly. However, the larger the change is, the more significantly the change is translated into changes in the vector representations of the faces, specifically- the predicted facial expression change (7-dimensional vector) or to the face 128-dimensional embedding vector change or to the most dominant expression.
 For example, 100\% natural faces are represented by the vector $(0,0,0,0,0,0,1)$ and 100\% surprised faces are represented by the vector $(0,0,0,0,1,0)$. The quick change from natural face to surprised face will be reflected as a large distance between the vectors.
A small change from a fully neutral face to a somewhat laughing face (for instance, $(0,0,0,0.1,0,0,0.9)$) is reflected by a small distance between the corresponding face expressions vectors. Regular conversations are characterized by a flow of small changes in vector representation, while significant changes indicate dramatic occurrences in the conversation, potentially anomalous ones.  We hypothesized that there would be small changes between facial expressions in a normal conversation. On the other hand, when an unusual event occurs, we expect many more or larger transitions to occur between subsequent facial expressions. 
The results of different techniques for detecting changes in time series can indicate a drastic change in the face due to an unusual event (see Figure~\ref{fig:maxdf} for illustration).

We constructed three time-series anomaly detection model variants and compared their performance.

The first model we used for anomaly detection is the \textit{Statistical Profiling Approach}~\cite{Rousseeuw2018AnomalyDB}. We considered and compared different numbers of frames to be used as the time window for calculating the rolling average and standard deviation. Points that deviate from the moving average by more than a pre-defined threshold relative to the moving standard deviation are considered an anomaly. \begin{equation}\label{equation_statistical_detection}
 \frac{actual\_value}{rolling\_average_{w}} > anomaly\_threshold \cdot rolling\_std\_dev_{w}
\end{equation}

Where $rolling\_average$ and $rolling\_std\_dev$ represent, respectively, the moving average and the moving standard deviation of a window of $w$ frames before the measured value. The parameter $w$ was tested with different values to get the optimal value (see Section~\ref{Experimental_Setup}).

The second method for anomaly detection is based on the \textit{Autoregressive Integrated Moving Average model (ARIMA)}~\cite{ZHANG2003159}. This methodology for predictive linear time series data is commonly used to predict expected future samples in a specific time frame, given the near history of the samples. We assume that if there is a significant difference between the predicted value and the actual observation - this can indicate an anomaly.  
The model can be represented as a function of three parameters: $p$, $d$, $q$, where  $p$ is the number of lag observations included in the regression model and
$d$ is the number of times that the raw observations are differenced, equivalent to the minimum number of differences needed to make the series stationary. In case the time series is stationary $d=0$,
$q$ represents the order of the moving average window and refers to the number of lagged forecast errors that should go into the ARIMA Model. In the context of our problem, $p$ is chosen based on the correlation between the training samples. If two samples are far away from each other, the correlation decreases. Thus,
the parameter $p$  should be chosen such that older samples that still have a high correlation to the next predicted value will contribute to the prediction. In addition, we took a small $p$ to keep the model effective and easy to train. 
Given the time series of embedding vector change, we trained a simple ARIMA model on the last 20-time series samples to predict the next sample. With this approach, samples will be considered as anomalous if the predicted value of the absolute vector change produces significant error relative to the actual value of the absolute vector change:
\begin{equation}
 \frac{predicted\_value - actual\_value}{actual\_value} > anomaly\_threshold.
\end{equation}

Where $predicted\_value$ is the ARIMA prediction model output, and  \\
\textit{anomaly\_threshold} is a pre-defined threshold. 

The third approach we used for anomaly detection is to identify and track dominant facial expressions change rather than the participant's embedding vector used in the first two methods. When nothing unusual happens, the dominant facial expression tends to remain constant or change only little along subsequent frames, with an occasional false classification of a different facial expression. On the other hand, special events are characterized by sharp changes in facial expressions and can include multiple transitions between different facial expressions throughout a short time window of frames. We attempted to identify time windows in which multiple expression transitions appeared in most of the frames, and we identified these windows as an anomaly in the participant's behavior. 
At the end of this anomaly detection step, we indicate for each participant separately and, according to each of the three methods, time samples that are likely to be anomalous, yet they are limited to indicating uncharacteristic changes in the embedding vector of single participants only and may not represent a general anomaly in the full meeting. 

\begin{figure*}[htbp]
\includegraphics[width=\textwidth]{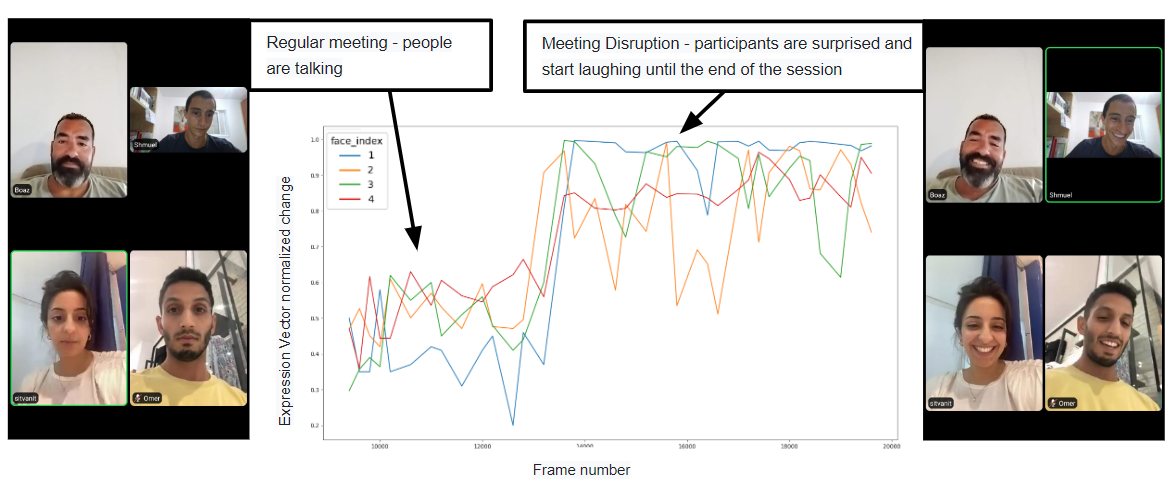}
\centering
\caption{Change in Expression Vector over Time: the left image is a snapshot from the beginning of the meeting when the four participants speak normally, and the change is relatively small. The right image is a snapshot taken from the continuation of the conversation when one of the participants did something foolish and all the other participants started laughing. The expression vector change is depicted by a single plot per participant. It is clear that the change in the vector representation of all four participants' expressions abruptly increases the time series graph around frame number 13,000 (Screenshot taken from the first author's Zoom call).}
\label{fig:fer_change_demo}
\end{figure*}

In Figure~\ref{fig:fer_change_demo}, we illustrate how a significant change in a video conference is seen and detected with the demonstration of the expression vector change over time.

 \subsection{Group Aggregations of  Anomalous Detections}
The fifth and final step of our pipeline is responsible for aggregating anomalies across individual participants into clusters of abnormal events that affect all meeting participants. This step can be referred to as a problem of aggregating anomalous samples of different faces into one cluster to account for an anomaly that co-occurs in multiple participants at the group level. 
While different techniques of clustering can be considered to tackle this problem, the challenge is to link anomalies in participants' faces at different time frames because participants may have their own reaction times and response characteristics with respect to the time of the unusual event. 
An empirical eyeball inspection of our data suggested that response duration periods can also range from a few seconds to tens of seconds, as participants continue to laugh, talk to each other, etc. Thus, we need to take into account the existence of the long and diverse time windows in which the unusual behavior of the participants could be noticed.

To identify clusters of anomalies, we chose to use the Density Based Spatial Clustering of Applications with Noise (DBSCAN) algorithm \cite{osti_421283}, which is very efficient at finding high-density time windows in the time axis where a relatively large portion of the participants show anomalous behavior at once. 
There are two key parameters of DBSCAN: 
\begin{enumerate}
\item \textbf{The distance that specifies the neighborhoods: }
Two points are considered to be neighbors in the same cluster if the distance between them is less than or equal to $\epsilon$. In our case, the $\epsilon$ parameter represents the maximal time delay between two anomalous points of two faces, or in other words, the typical difference between response times of different individuals to the same event. We tested a range of values between 5 and 21 frames of the epsilon parameter to obtain the optimal value that would maximize performance, as shown in Section~\ref{Experimental_Setup}. 
\item \textbf{The minimum number of data points to define a cluster:}
In our case, this is the minimal number of participants with anomalous behavior we require to declare a group meeting anomaly event. This number should increase in proportion to the number of participants in the conversation, and practically, this parameter helps us to reduce the fraction of False Negative detections of an anomaly event. The probability that such a change will occur to numerous participants in a small time window decreases as the number of participants increases unless there is, in fact, an unusual group disruptive event. We defined a required minimal $ratio\ 
 of\  participants$ to detect an anomaly. This parameter expresses the ratio between the number of participants with abnormal behavior and the total number of participants in the meeting. 
Multiplying the number of participants by the $ratio\ of\ participants$ gives the threshold number of participants to identify an anomaly in the conference.
\end{enumerate}
\begin{figure}[h]
    \includegraphics[width=0.45\textwidth]{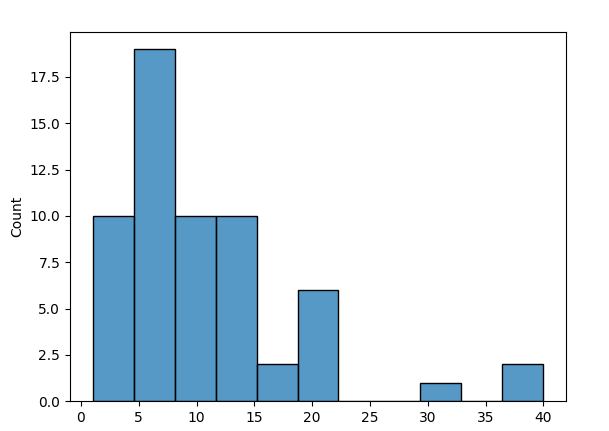}
    \centering
    \caption{Histogram of anomalous events time windows [sec].}
    \label{fig:anomalies_times_hist}
\end{figure}

\section{Experiments and Results}\label{experiments_results}
\subsection{Dataset}
\begin{figure}[h]
    \includegraphics[width=0.45\textwidth]{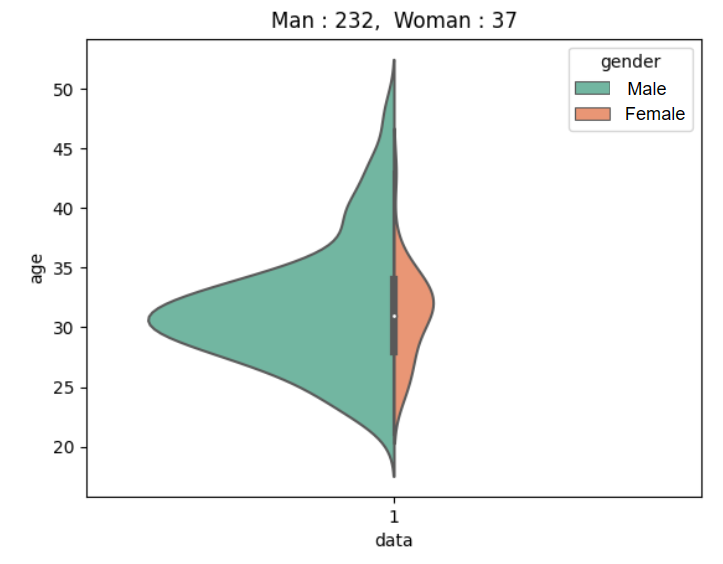}
    \centering
    \caption{Age and Gender Distribution of Meeting Participants  in the Collected Dataset.}
    \label{fig:age_gender_dist}
\end{figure}

Due to the lack of labeled video conferencing video datasets in general
and abnormal events detection specifically, we manually create our in-house  first-of-its-kind dataset of VC recordings containing annotated data of anomalies and unusual events, which we offer for public use. Data is collected from
video recordings of meetings containing a zoombombing event or an unusual event during the meeting. 
We collected thirty videos with a total length of more than 8 hours and 65 abnormal events with various time windows, as shown in Figure~\ref{fig:anomalies_times_hist}. 
The number of participants in the dataset ranges from 4 to 25, with more dominant representation for younger ages, around the age of 30, and for men, as can be seen in Figure~\ref{fig:age_gender_dist}.

In each video, to define an abnormal event, we first identified the start and end time points of meeting disruptions  and manually tagged them for algorithm evaluation.
The abnormal time windows were characterized by the response of other participants to the disruption. We considered abnormal events, such as contagious
laughter or mass panic, or another change that disrupted the ordinary course of the  conversation. In a calm and relaxed conversation, a light chuckle may be considered an abnormal event, while in a more dynamic meeting that contains many giggles, only a deep and unusual laugh will be recognized as an abnormal event.\footnote{For example: https://www.youtube.com/embed/48KjeTeWe7Y}

\subsection{Experimental Setup}\label{Experimental_Setup}

The Parameter choices and evaluation of individual steps of our pipeline are described below.

\begin{enumerate}
 \item \textbf{Face Detection in Video Conferencing Images.}
 To evaluate the quality of our dual model-based detection method for application in the domain of video conferencing, we randomly selected 100 video conferencing collage images and applied the detection method. The performance reached a recall of 87\% and a precision of 95\%. These values were obtained when we set an 0.8 Intersection Over Union (IOU) threshold between the YOLO model and the RetinaFace model.

\item \textbf{Faces Tracking.}
We manually inspected dozens of video conference videos and noticed that in some cases, the faces’ locations in the faces grid change or get shuffled, but usually, the grid remains constant.
We examined the IOU between the location of the participants in consecutive frames. We observed that in 95\% of the cases, the IOU is higher than 0.5, meaning that there are only slight movements of the participant's face.
As explained in Section~\ref{detection_method}, we utilized the bounding box achieved in the faces detection stage to associate a specific participant's face in consecutive frames. We measured the IOU between each pair of faces in the two successive frames. We associated the faces pairs with each other if all pairs in the current and previous image had IOU greater than 0.5. Otherwise, we linked the current frame faces to the last frame faces by comparing the detected faces in each frame to the faces in the previous frame with the face recognition tool.
To examine the performance of this step, we manually evaluated the tracking ability in ten 5-minute slices of videos containing a total of 75 participants. In 70  of the 75 participants ($93.33\%$), we verified the correct continuous tracking of each participant throughout the video. Four participants' tracks (in four different videos) were split twice, and one participant's track (in another video) was split five times.
In the case of lost tracking, the algorithm established a new participant even though he was a known active participant that the algorithm failed to identify as the same one. In these cases, the algorithms continued to work regularly, looking for anomalies in the "new" participant and refreshing the search from that time point.

\item \textbf{Features Extraction} 
 We used the  3589 validation images from the FER2013  dataset to evaluate expression classification models. We achieved a precision of 60\% using the trained FER model.

In Figure~\ref{fig:zoom4faces}, we illustrate the output of the first three steps: faces have been detected and tracked (id is given according to the previous frame), and expression is extracted.

\begin{figure}[t]
\includegraphics[scale=0.3]{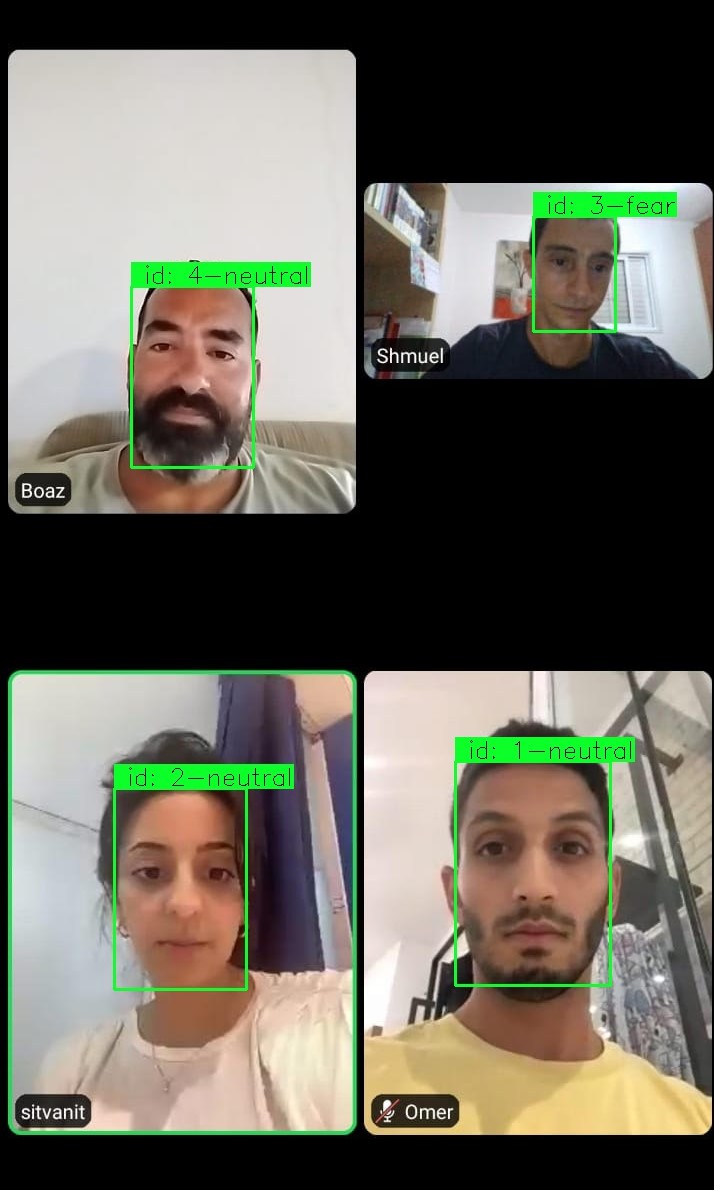}
\centering
\caption{Face Detection, Tracking, and Expression Extraction in Video Meetings (Screenshot from the first author's Zoom call).}
\label{fig:zoom4faces}
\end{figure}

\item \textbf{Change Detection.}\label{change_detection} 
For evaluation purpose, we used a k-fold cross-validation technique. In cross-validation, the data are randomly partitioned into k subsets or folds and the predictive model is trained k times~\cite{Refaeilzadeh2009}. At each training stage, one fold was used as the test set, and the remaining k-1 folds were used as a training set. The k-fold cross-validation provides a better generalization of the model~\cite{ISHIBUCHI201322}: The repetitions of tests by cross-validation ensure that the prediction errors are not randomly good or bad and are an average of multiple runs. In the next steps and in evaluating the final results, we have implemented a 10-fold cross-validation approach. 
\textit{Statistical Profiling Approach:} We used frame windows with a size of 5 to 15 frames in increments of 2 to detect anomalous samples. As a threshold for setting an anomaly detection, we tested thresholds of 1.5 to 3 standard deviations in increments of 0.1.
The optimal performance was achieved with a moving average with a window size of 7 frames and a threshold of 1.8 STD relative to the moving average. Figure~\ref{fig:anomaly_with_ma3} demonstrates how anomalies are detected with a moving average as explained in Equation~\ref{equation_statistical_detection}. 

\begin{figure*}[t]
    \includegraphics[width=0.8\textwidth]{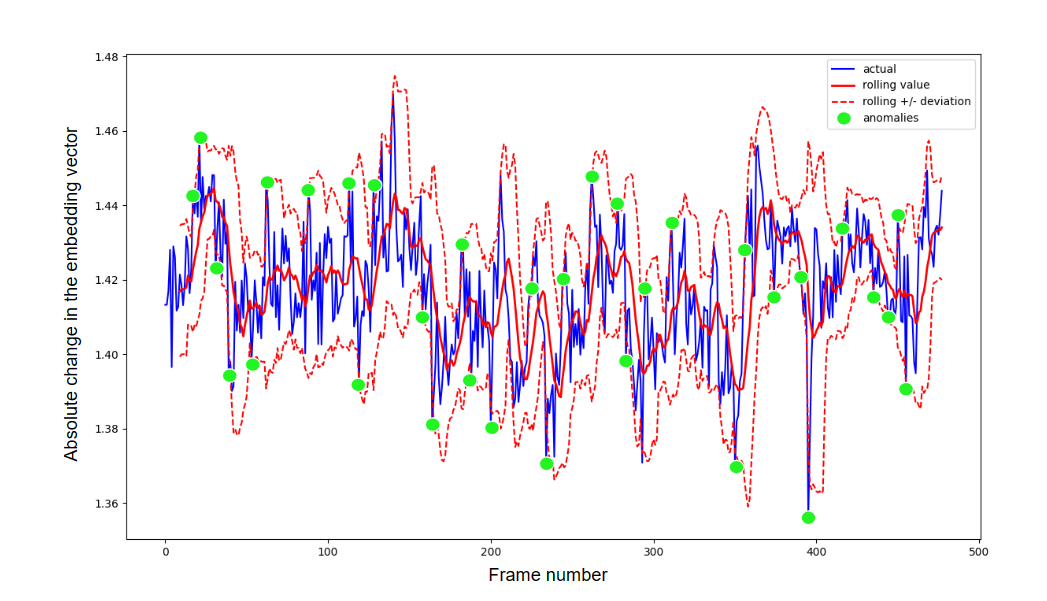}
    \centering
    \caption{Example of simple anomaly detection with Statistical Profiling Approach on one of the dataset videos. Many points are detected in this case as anomalies because of a significant events that occurred at the meeting and was reflected in the participant's response.}
    \label{fig:anomaly_with_ma3}
\end{figure*}

\textit{ARIMA:} To perform the next sample prediction in the time series, we used the Python package $pmdarima$.\footnote{\url{https://pypi.org/project/pmdarima/##description}} To find the optimal values for the parameters $p$, $d$, and $q$, we used the automatic $auto-arima$ tool, which performs a grid search to find the optimal values. After the optimization, we used the ARIMA model with parameter values of $p=2$, $d=0$, and $q=2$. 
In Figure~\ref{fig:prediction_with_arima}, we illustrate the time series absolute change prediction using the ARIMA model, where each predicted sample is a prediction result based on the 20 preceding samples, as explained in Section~\ref{change_detection_section}.
\begin{figure*}[t]
    \includegraphics[width=0.8\textwidth]{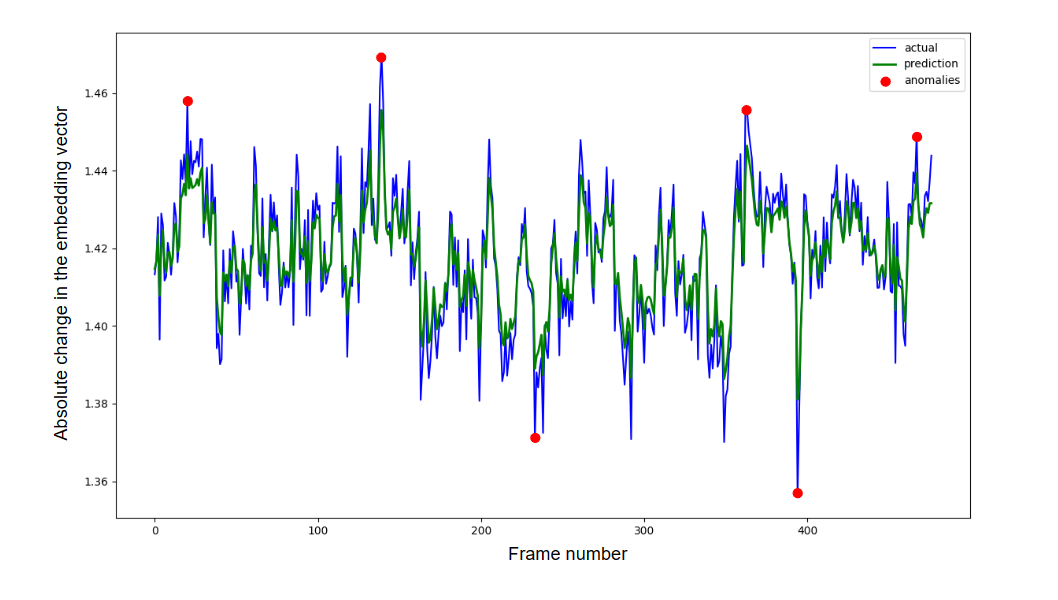}
    \centering
    \caption{Prediction of face embedding vector change with ARIMA. Abnormal points are marked by red dots.}
    \label{fig:prediction_with_arima}
\end{figure*}

Next, we applied ARIMA model on a sliding window of each participant time series to compute the predicted value based on the  20-preceding-frames window.  

\textit{Facial expressions change:} In the third method for identifying areas saturated with such expressions changes (\ref{third_approach_detection}), we examined different sizes of time windows from 3  to 9 frames with 2-frame increments. We consider a time window as a suspect for anomaly if  most participants' faces within the window exhibit transition between different facial expressions. 

\item \textbf{Accumulated change.}\label{epsilon_search} To optimize the DBSCAN algorithm, we compared the results obtained with different values of $\epsilon$  between 5 and 21 frames with 2-frame increments. The best performances over the final precision were obtained with the value $\epsilon = 9$.

\end{enumerate}

\subsection{Evaluation Metrics}

We have considered anomalous events as drastic change that begins in the participants' behavior relative to their behavior in the previous frames. 
If a significant change occurs and persists over several consecutive frames, the event is defined as a continuous event over time. A sequence of frames without substantial changes in the participants'  behavior is defined as a regular event.
A True Positive - $TP$ event is defined for each frame separately if a significant change was correctly detected, and a False Positive- $FP$ event is defined as an event in which a regular occurrence has been identified as an anomaly.
Similarly, a True negative - $TN$ event is defined for each frame if no significant event was correctly detected as a regular event, and a False Negative - $FN$ event is defined as an actual anomaly event incorrectly classified as a regular event.

Finally, given the definitions for $TP$,  $FP$, $TN$, and $FN$, we  define the four metrics with that we used to evaluate the algorithm's performance, depending on different parameters.
Recall is used to evaluate how complete the result is, and Precision is used to evaluate the accurracy of the result. Specificity and FPR represent the symmetric picture for the negative results.\footnote{In many cases, anomaly detection algorithms use the ROC-CURVE metric for evaluation. We do not use this metric because the nature of our algorithm generates binary classifications without a probability score, so we only have access to a single point in the ROC operating range.}

\subsection{Anomaly Detection Results}
To summarize the performance of the complete pipeline, we first evaluated
the performance
of our algorithm by comparing the performance of different combinations of configurations (see Table \ref{table:ta_res}). We tested the performance of each of the pipeline steps with different parameter choices and presented the average performance of the 10-folds test as described in section~\ref{Experimental_Setup} in terms of Recall and Precision.
\begin{table*}[h]
\centering
\caption{Interruptions Detection Results}
{\begin{tabular}{ |c|c|c|c|c|c| }
 \hline
 \multicolumn{2}{|c|}{\textbf{Method}} & \multicolumn{4}{|c|}{\textbf{Results}}\\
 \hline
 \textbf{Feature extraction}  & \textbf{Change detection} & \textbf{Recall } & \textbf{Precision} & \textbf{TNR} & \textbf{FPR} \\
 \hline
{embedding 128-dim vector} 
& Arima & 50.1\% & 83.2\%   &  98.1\%& 0.9\% \\
& Statistical profiling & 42.9\% & 75.0\%   &  97.8\%& 1.4\% \\
\hline
{expressions 7-dim vector}
& Arima & 71.6\% & 90.5\%   & \textbf{99.2\%} & \textbf{0.5\%} \\
& Statistical profiling & \textbf{72.5\%} & \textbf{92.3\%}   &  99.1\%& 0.8\% \\
\hline
deep-face expression & Majority of changes & 67.1\% & 84.8\%  &  97.7\% & 2.6\% \\[1ex] 
\hline
\end{tabular}}
\label{table:ta_res}
\end{table*}

The best Recall of 72.5\% was obtained using the facial expressions' embedding vector and statistical profiling to detect anomalies in the participant's behavior. The best performance in terms of Precision was 92.3\%, also obtained by using facial expressions' embedding vector and statistical profiling. 
Using an ARIMA-based approach with facial expressions embedding produces similar results to the statistical profiling approach with slightly improved TNR and FPR.
The difference between finding the anomaly based on the statistically-based approach versus the ARIMA-based approach provided similar results, with a slight preference in favor of the statistical profiling approach.

\begin{figure}[h]
\centering
\includegraphics[scale=0.3]{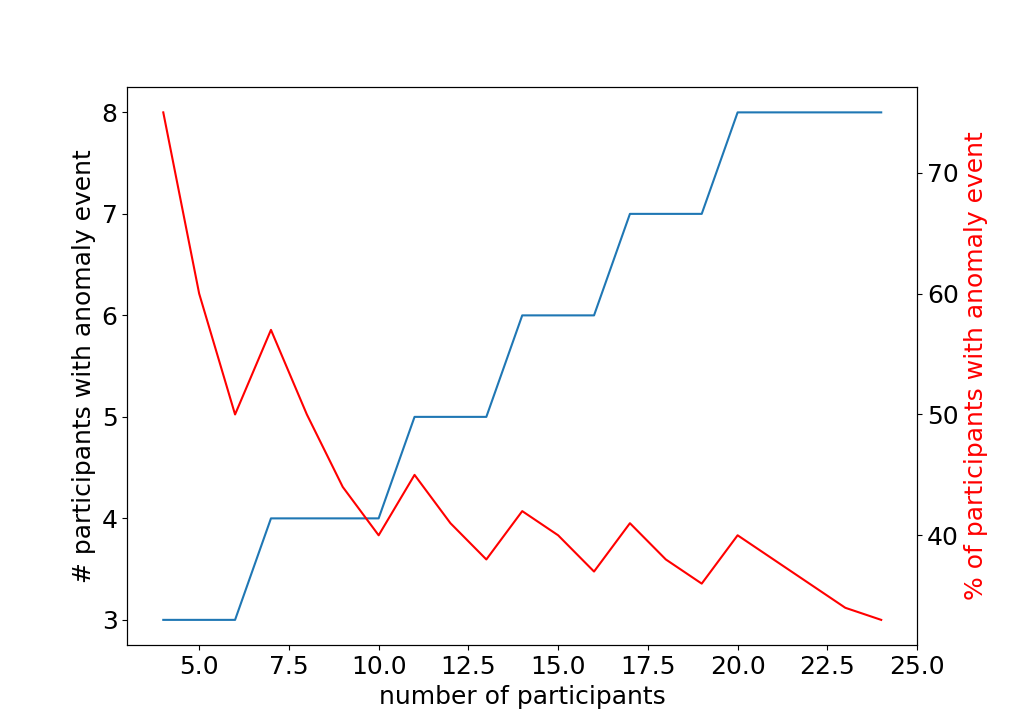}
\centering
\caption{The required number of participants with anomalous behavior to detect anomaly events in the VC (blue left Y axis), and their percentage of total participants (red right Y axis).}
\label{fig:participants_ratio}
\end{figure}

The number of participants in whom abnormal behavior needs to be detected to identify a group anomaly increases as the number of participants in the meeting increases. However, as the meeting size increase, it becomes easier to detect abnormal events because we have more chances to detect a group of abnormal participants: The proportion between the number of abnormal participants and the total number of participants in the conversation has a decreasing trend (see Figure~\ref{fig:participants_ratio}).

\section{Discussion}\label{discussion}

We propose a novel method that could be widely used for an automatic detection of abnormal events in video conferencing videos.
Hundreds of millions of users with various use cases utilize video conferencing on a daily basis. 
This method can be applied to annotate important time frames in online meetings and is potentially practical for detecting and dealing with cyberbullying attacks in video conferences. Moreover, it could be used for educational purposes helping to highlight important points in the video conferencing recordings.

The results show features that represent facial expressions can be used to detect anomalies in video meetings. Even though there is a disagreement in the research community about whether emotion detection algorithms detect the real emotion of the subject, they still detect changes in expression.
Our results demonstrate the synchronous change in facial expression of a high percentage of the participants in a meeting shows that something unusual is happening.
Such an event may be a zoombombig or another cyberbullying attack.
Also, it can detect a variety of other types of events, such as someone telling a good joke, a manager announcing a major announcement, or even when the students are confused and do not understand the explanation.

We find that there are considerable differences between different features. We hypothesize that the identification of the anomalies using the 128-dimensional embedding vector was less suitable because this vector was not trained to detect subtle changes in the face that indicate a change in conversation but was trained in general to differentiate between different people and the deep-face expression features were too noisy and therefore produced inferior results. 
In terms of change detection methods, there was almost no difference between ARIMA and statistical profiling. 
However, ARIMA had consistently lower false positive rates, which makes it a better suit for cases where false positive is less tolerated.

Additionally, we have shown that the proposed method can be used as a whole to detect unusual events in video calls with good performance in terms of both recall and precision, along with low rates of false alarms. Accurate detection of such events can assist video conference organizers in monitoring the meeting and making sure that no one interrupts the conversation. Unusual events in the meeting may be flagged on time and addressed accordingly. This ability may help prevent VC disruption and increase meetings' effectiveness.

The algorithm presented in this work is prone to several limitations.
First, the dataset we collected may be biased toward a specific use case. The data harvested from publicly available YouTube videos containing ZOOM platform VC recordings only and therefore does not contain data from meetings that were not publicly published online or meetings from other VC platforms. In addition, it is based on a relatively small dataset that likely does not represent a large enough variety of participants and meeting culture. Moreover, because the datatset is small, the annotated anomalies in the dataset also represent a relatively small variety of interruptions and situations that can occur in the real world. All of these may make the current state of the algorithm overfitted to the specific dataset in use, thus it may fail in generalizing for cases in which the meeting characteristics are fundamentally different from the characteristics of the data. Future work is needed to expand and enrich the dataset to contain a greater variety of participants and events.

Second, in terms of parameters in use, the results presented here reflect the use of default parameters calibrated for the entire dataset, and while the proof-of-concept results are satisfying and promising, additional parameter calibration can be performed for each video separately to improve the ability to detect anomalies. For example, the parameters $p$, $q$, and $d$ of the ARIMA model or the statistical profiling parameters like $window\_size$ and $std\_threshold$.

The future goals of our research span several directions:
First, we would like to test methods that can improve the method performance.
For instance, LSTM and Spatiotemporal Autoencoder can be used for anomaly detection~\cite{7344872}~\cite{Chong2017AbnormalED}, and enhanced versions of YOLO can be used for face detection~\cite{wang2022yolov7}.
Also, we plan to adapt the pipeline to perform in face-to-face conversations where the participants do not look directly at the camera and are in a less fixed position, and adjust the pipeline to work in real-time, so it will be possible to automatically monitor and detect the anomalies at the moment they occur and not only in an offline mode. This way, the algorithm will work automatically and in flexible scenarios.

\section{Conclusions}\label{conclusions}
Today's post-Covid world experiences a significant increase in video conferencing as a daily means of conducting meetings in various fields, as businesses and other organizations adopt remote working, social and education arrangements. Along with the surge in video conference use comes the recognition of the importance of having safe and effective meetings without interruptions and disruptions. 
However, to the best of our knowledge, no available open methods are tailored for monitoring and detecting abnormal events in video conferencing meetings. In this work, we present the first method to automatically identify abnormal events in streams of video meetings, aiming to capture event-related interruptions to a conversation. Additionally, we curated the first dataset of video conference recordings containing various annotated unusual and anomalous VC events.

We demonstrated that users' emotions, as deduced from their facial expressions, have an important potential for detecting abnormal events in video meetings. Identifying that multiple participants' expressions have changed synchronously is already sufficient to conclude that an interesting event has occurred. Our method can be applied for automatically screening any video conferencing input and can be fine-tuned to suit specific domains for an even improved performance. In a world where more and more activities transfer to VC, tools that detect interesting and abnormal events are of majorly increased importance.

\section*{Code and Data Availability}\label{opencode_and_dataset}
The open-code framework is available at
\url{https://github.com/shmuelhorowitz/Interruptions_detection}.

Due to the private nature of the video conference data, there is a process of requesting publication permission from all video owners to publish the videos publicly. After receiving the approval, it will be possible to obtain the dataset upon request from the corresponding author.

\clearpage

\phantomsection
\bibliographystyle{unsrt}
\bibliography{sample}


\end{document}